%% file: main.tex
\documentclass{article}

\usepackage{xfp}
\usepackage{url}
\usepackage{hyperref}

\usepackage{iclr2027_conference,times}
\iclrfinalcopy
\usepackage{caption}
\input{macros}

\input{numbers/pids0910}  

\title{Seeing Depth but Choosing Cues in Vision-Language Models}

\author{
Zhengtao Zou \\
Aalto University \\
\texttt{zhengtao.zou@aalto.fi}
\And
Pekka Marttinen \\
Aalto University \\
\texttt{pekka.marttinen@aalto.fi}
}

\begin{document}
\maketitle

\fancyhead{}
\renewcommand{\headrulewidth}{0pt}

\input{sections_0907/abstract}
\input{sections_0907/intro}
\input{sections_0907/related_work}

\input{sections_0907/analysis}
\input{sections_0907/probing}
\input{sections_0907/causal}

\input{sections_0907/limitations}
\label{sixend}

\label{maintextend}

\bibliographystyle{iclr2027_conference}
\bibliography{references}

\appendix
\input{backmatter/ai_statement}
\input{backmatter/repro_statement}
\input{appendix/c_external}
\input{appendix/h_benchmark_construction}
\input{appendix/i_stimuli}
\input{appendix/a_protocols}

\end{document}

%% file: macros.tex
\usepackage{graphicx}
\usepackage{booktabs}
\usepackage{amsmath}
\usepackage{amssymb}
\usepackage{subcaption}
\usepackage{xcolor}
\usepackage{multirow}
\usepackage{microtype}

\newcommand{\qeight}{Qwen3-VL-8B}

\newcommand{\gtwelve}{Gemma-4-12B}


%% file: numbers/pids0910.tex
\makeatletter
\newcommand{\pidset}[2]{\expandafter\gdef\csname pid@#1\endcsname{#2}}
\newcommand{\pid}[1]{%
  \ifcsname pid@#1\endcsname
    \csname pid@#1\endcsname
  \else
    \PackageError{pids}{Undefined PID #1 -- not in the provenance registry}{}%
  \fi}
\newcommand{\pidciset}[2]{\expandafter\gdef\csname pidci@#1\endcsname{#2}}
\newcommand{\pidci}[1]{%
  \ifcsname pidci@#1\endcsname
    \csname pidci@#1\endcsname
  \else
    \PackageError{pids}{Undefined PID interval #1 -- no registered bootstrap bounds}{}%
  \fi}
\makeatother
\newcommand{\pidpct}[1]{\pid{#1}\%}

\pidset{100}{89.4}  
\pidset{111}{59.8}  
\pidset{121}{42.4}  
\pidset{128}{75.0}  
\pidset{150}{88.2}  
\pidset{151}{74.2}  
\pidset{152}{68.8}  
\pidset{154}{89.2}  
\pidset{155}{77.5}  
\pidset{156}{75.0}  
\pidset{157}{77.5}  
\pidset{158}{89.2}  
\pidset{159}{75.6}  
\pidset{160}{69.4}  
\pidset{162}{85.2}  
\pidset{163}{60.3}  
\pidset{164}{39.6}  
\pidset{166}{86.9}  
\pidset{167}{64.1}  
\pidset{168}{52.8}  
\pidset{170}{86.2}  
\pidset{171}{53.1}  
\pidset{172}{46.5}  
\pidset{174}{89.5}  
\pidset{175}{66.5}  
\pidset{176}{68.8}  
\pidset{200}{1.000}  
\pidset{209}{0.862}  
\pidset{239}{0.998}  
\pidset{289}{52.4}  
\pidset{305}{53.6}  
\pidset{400}{1.00}  
\pidset{432}{100.0}  
\pidset{448}{89.2}  
\pidset{452}{78.9}  
\pidset{453}{80.6}  
\pidset{454}{87.9}  
\pidset{463}{76.1}  
\pidset{464}{77.8}  
\pidset{465}{86.2}  
\pidset{469}{52.6}  
\pidset{470}{38.2}  
\pidset{471}{78.4}  
\pidset{475}{57.4}  
\pidset{476}{47.2}  
\pidset{477}{78.0}  
\pidset{481}{65.1}  
\pidset{482}{61.8}  
\pidset{483}{85.2}  
\pidset{487}{72.7}  
\pidset{488}{73.6}  
\pidset{489}{84.9}  
\pidset{500}{88.2}  
\pidset{516}{77.5}  
\pidset{517}{75.0}  
\pidset{518}{84.9}  
\pidset{556}{87.9}  
\pidset{557}{76.1}  
\pidset{558}{72.2}  
\pidset{590}{88.9}  
\pidset{592}{69.2}  
\pidset{632}{95.7}  
\pidset{633}{93.5}  
\pidset{638}{95.5}  
\pidset{639}{93.6}  
\pidset{642}{0.000753}  
\pidset{644}{85.9}  
\pidset{645}{85.9}  
\pidset{658}{84.6}  
\pidset{852}{0.574}  
\pidset{947}{0.42}  
\pidset{976}{0.710}  
\pidset{1076}{89.5}  
\pidset{1108}{99.6}  
\pidset{1180}{84.6}  
\pidset{1181}{66.5}  
\pidset{1182}{54.9}  
\pidset{1188}{87.5}  
\pidset{1189}{70.8}  
\pidset{1190}{63.2}  
\pidset{1249}{74.2}  
\pidset{1250}{68.8}  
\pidset{1251}{87.9}  
\pidset{1338}{68.4}  
\pidset{1339}{53.5}  
\pidset{1340}{84.6}  
\pidset{1341}{70.8}  
\pidset{1342}{63.2}  
\pidset{1343}{87.5}  
\pidset{1442}{0.219}  
\pidset{1443}{0.382}  
\pidset{1444}{0.326}  
\pidset{1446}{0.446}  
\pidset{1447}{-0.155}  
\pidset{1448}{0.022}  
\pidset{1450}{0.015}  
\pidset{1451}{0.541}  
\pidset{1452}{0.548}  
\pidset{1454}{-0.030}  
\pidset{1455}{0.349}  
\pidset{1456}{0.593}  
\pidset{1500}{0.020}  
\pidset{1858}{95.4}  
\pidset{1859}{54.8}  
\pidset{1860}{34.8}  
\pidset{1862}{94.4}  
\pidset{1863}{50.2}  
\pidset{1864}{42.8}  
\pidset{1866}{97.2}  
\pidset{1867}{34.6}  
\pidset{1868}{45.0}  
\pidset{1870}{98.6}  
\pidset{1871}{42.0}  
\pidset{1872}{7.4}  
\pidset{1874}{97.6}  
\pidset{1875}{43.8}  
\pidset{1876}{48.8}  
\pidset{1878}{93.6}  
\pidset{1879}{44.2}  
\pidset{1880}{23.2}  
\pidset{1882}{98.4}  
\pidset{1883}{39.0}  
\pidset{1884}{24.2}  
\pidset{1886}{98.4}  
\pidset{1887}{43.2}  
\pidset{1888}{34.0}  
\pidset{1890}{94.6}  
\pidset{1891}{32.6}  
\pidset{1892}{46.8}  
\pidset{1931}{44.8}  
\pidset{1932}{32.2}  
\pidset{1933}{97.8}  
\pidset{1937}{53.4}  
\pidset{1938}{57.0}  
\pidset{1939}{98.6}  
\pidset{2012}{41.8}  
\pidset{2013}{41.4}  
\pidset{2014}{45.6}  
\pidset{2015}{36.0}  
\pidset{2016}{42.2}  
\pidset{2017}{35.2}  
\pidset{2018}{46.6}  
\pidset{2019}{38.4}  
\pidset{2020}{49.2}  
\pidset{2021}{43.4}  
\pidset{2022}{48.6}  
\pidset{2023}{51.8}  
\pidset{2024}{63.4}  
\pidset{2025}{53.2}  
\pidset{2026}{51.0}  
\pidset{2027}{42.6}  
\pidset{2032}{0.19}  
\pidset{2033}{0.69}  
\pidset{2034}{0.43}  
\pidset{2035}{1.02}  
\pidset{2036}{-0.19}  
\pidset{2037}{0.34}  
\pidset{2038}{0.13}  
\pidset{2039}{0.68}  
\pidset{2040}{0.83}  
\pidset{2041}{0.20}  
\pidset{2042}{0.29}  
\pidset{2043}{0.65}  
\pidset{2157}{99.8}  
\pidset{2158}{88.8}  
\pidset{2159}{75.4}  
\pidset{2161}{99.6}  
\pidset{2162}{90.8}  
\pidset{2163}{71.4}  
\pidset{2165}{100.0}  
\pidset{2166}{85.0}  
\pidset{2167}{84.4}  
\pidset{2169}{96.8}  
\pidset{2170}{83.6}  
\pidset{2171}{47.0}  
\pidset{2173}{99.8}  
\pidset{2174}{77.6}  
\pidset{2175}{77.4}  
\pidset{2177}{98.0}  
\pidset{2178}{73.2}  
\pidset{2179}{48.6}  
\pidset{2185}{96.8}  
\pidset{2186}{51.8}  
\pidset{2187}{37.6}  
\pidset{2211}{93.0}  
\pidset{2212}{91.0}  
\pidset{2213}{94.0}  
\pidset{2214}{93.8}  
\pidset{2215}{94.0}  
\pidset{2216}{88.8}  
\pidset{2260}{89.4}  
\pidset{2261}{89.8}  
\pidset{2262}{94.8}  
\pidset{2263}{94.4}  
\pidset{2267}{66.6}  
\pidset{2268}{57.4}  
\pidset{2269}{88.8}  
\pidset{2273}{51.6}  
\pidset{2274}{41.5}  
\pidset{2275}{94.6}  
\pidset{2282}{51.0}  
\pidset{2283}{47.6}  
\pidset{2284}{89.8}  
\pidset{2343}{78.9}  
\pidset{2344}{80.6}  
\pidset{2352}{75.5}  
\pidset{2353}{73.7}  
\pidset{2354}{87.7}  
\pidset{2361}{77.0}  
\pidset{2362}{77.8}  
\pidset{2363}{90.2}  
\pidciset{1442}{[0.151, 0.288]}  
\pidciset{1443}{[0.301, 0.460]}  
\pidciset{1444}{[0.258, 0.402]}  
\pidciset{1446}{[0.360, 0.534]}  
\pidciset{1447}{[-0.233, -0.083]}  
\pidciset{1448}{[-0.055, 0.098]}  
\pidciset{1450}{[-0.036, 0.066]}  
\pidciset{1451}{[0.484, 0.603]}  
\pidciset{1452}{[0.482, 0.621]}  
\pidciset{1454}{[-0.079, 0.020]}  
\pidciset{1455}{[0.287, 0.412]}  
\pidciset{1456}{[0.525, 0.660]}  
\pidciset{2032}{[0.12, 0.26]}  
\pidciset{2033}{[0.60, 0.78]}  
\pidciset{2034}{[0.35, 0.51]}  
\pidciset{2035}{[0.90, 1.16]}  
\pidciset{2036}{[-0.27, -0.11]}  
\pidciset{2037}{[0.26, 0.41]}  
\pidciset{2038}{[0.08, 0.18]}  
\pidciset{2039}{[0.61, 0.74]}  
\pidciset{2040}{[0.75, 0.93]}  
\pidciset{2041}{[0.16, 0.25]}  
\pidciset{2042}{[0.24, 0.34]}  
\pidciset{2043}{[0.58, 0.73]}  


%% file: sections_0907/abstract.tex
\begin{abstract}
Do vision-language models use depth information to judge which object is closer, or rely on image-plane cues? We investigate this question through controlled depth--cue conflicts, representation analysis, and interventions. Among the many image-plane cues relevant to depth perception, we focus on apparent size and vertical position. We compare depth judgments under cue-consistent and cue-conflict conditions in both rendered scenes and real-world photographs. Across model families and scales, depth judgments systematically shift toward misleading cues. Yet the correct depth order remains decodable from the queried objects' visual representations, including when the model answers incorrectly. Image and category controls support interpreting this signal as image-dependent depth information beyond the two target cues. Strengthening the model's own depth readout through object-level write-back can correct conflict decisions without updating model parameters or using ground-truth depth annotations. Interventions show that the effect of this correction depends on the direction and target tokens of the written signal, while its strength varies across architectures and domains. Together, these findings reveal a mismatch between a model's ability to represent depth and use this information when deciding the answer. VLMs can retain useful depth information while allowing competing image-plane cues to exert greater influence on their answers.
\end{abstract}

%% file: sections_0907/intro.tex
\section{Introduction}

Picture a scene of you sitting in a living room. A mug is on the table in front of you, and a sofa is across the table. In the real world, any three-year-old can tell that the mug is closer to you than the sofa. However, for state-of-the-art vision-language models (VLMs), that answer is less reliable, despite their strength elsewhere~\citep{blink}. This paper asks why this happens and answers with a diagnosis. Humans often use statistical shortcuts, or cues, about the front-behind relationship between objects. For example, nearer things tend to sit lower in the frame and look larger. We hypothesize that VLMs may do the same. These cues are genuinely predictive in natural 2D images and are directly available from the image plane. A learner rewarded on ordinary photographs therefore has every statistical reason to use them. We ask whether VLMs actually use depth information when judging depth relations, or instead rely on these simpler 2D cues.

To separate depth from projected-size and vertical-position cues, we construct two controlled diagnostic settings. The first uses rendered scenes in which size and vertical position are decorrelated from true depth. The second uses real-world photographs with sensor depth, from which we select cases where these 2D cues point to the wrong answer. Across eleven VLMs from five families, including multiple model scales and a checkpoint fine-tuned for spatial reasoning, performance drops sharply when the cues conflict with true depth. Neither model scaling nor spatial fine-tuning eliminates this pattern. When depth and 2D cues disagree, VLM behavior systematically follows the cues. Code is available anonymously.%
\footnote{\href{https://anonymous.4open.science/r/Seeing-Depth-but-Choosing-Cues-in-Vision-Language-Models-561C/README.md}{Anonymous code repository.}}

\begin{figure}[ht]
\centering
\includegraphics[width=\linewidth]{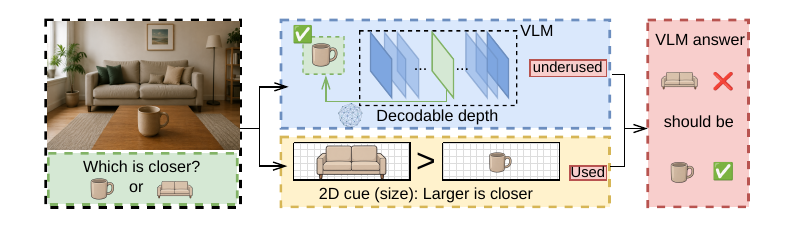}
\vspace{-20pt}
\caption{\textbf{Schematic illustration.} A conflict item. The sofa looks larger, so the size cue favors the
sofa. Yet the same forward pass contains the correct depth relation in its
visual representations, while the final decision follows the conflicting
cue.}
\label{fig:overview}
\end{figure}

Prior work has identified perspective shortcuts and spatial entanglement in VLMs~\citep{min2026far}, as well as steerable spatial directions~\citep{spatialid}. We show that correct depth order remains linearly decodable from the difference between the two objects' visual representations even when the model follows a conflicting cue. Decoding remains accurate when projected size and vertical position are decorrelated from depth, while image and category controls support its dependence on the current visual input.
We then strengthen this depth information through opposite updates to the two objects' representations, with the update sign determined by which object the probe predicts is closer. This object-level write-back requires neither evaluation depth labels nor updates to VLM parameters. It improves conflict accuracy across several model families, including on real photographs, although effects vary across architectures and domains. Together, these findings reveal a mismatch between available depth information and its influence on the answer, which object-level write-back can partially resolve.

Our contributions are:

\begin{itemize}
    \item We show across eleven VLMs from five families that depth judgments
    systematically follow image-plane size and vertical-position cues when
    those cues conflict with true depth.

    \item We identify a linearly decodable depth signal in visual object contrasts and show that it remains informative even when the model produces a cue-supported error.
    
    \item We show that object-level write-back, oriented by the model's own depth readout, can correct conflict decisions without updating VLM parameters. Matched controls test the specificity of this effect to the direction, target tokens, and write-back configuration.
\end{itemize}

%% file: sections_0907/related_work.tex
\section{Related Work}

\paragraph{Shortcuts in spatial reasoning.}
\citet{min2026far} show that VLM depth judgments are sensitive to
perspective cues, particularly vertical position.
We examine the internal basis of this behavior by controlling
apparent size and vertical position and testing whether depth
remains represented when answers follow conflicting cues.

\paragraph{Spatial representations and their influence.}
\citet{liao2026decodable} distinguish grounded, prior-driven, and
inverted depth representations through probing and visual ablations.
Our focus is the competition between image-dependent depth information
and misleading image-plane cues.
\citet{spatialid} identify linear spatial directions and show that
steering them can alter spatial judgments, including depth.
We instead read depth from queried-object contrasts and use
self-oriented write-back to test its influence under controlled
cue conflicts.
Following causal representation analyses
\citep{belrose2023leace,iti,caa}, our controls distinguish
decodability from intervention effects on the final answer.

\paragraph{Improving spatial reasoning.}
SpatialVLM~\citep{chen2024spatialvlm} and
SpatialRGPT~\citep{spatialrgpt} improve spatial competence through
additional supervision, while contrastive decoding methods modify
generation without updating model weights
\citep{li2023contrastive,leng2024mitigating}.
We use a frozen model to test how
available depth information influences decisions when it conflicts
with apparent size or vertical position.

%% file: sections_0907/analysis.tex
\section{Controlled Diagnosis of Depth--Cue Competition}
\label{sec:diagnosis}

\subsection{Experimental Setup}
We study pairwise depth judgments in which a model decides which of two objects is closer to the camera. Real-world images contain many depth-relevant cues, such as texture and shadow; we focus on apparent size and vertical position. Our rendered setting controls their agreement with true depth, while our natural-image setting selects corresponding conflicts using sensor-derived depth order. We vary the relation of these cues to depth to create three conditions. In the \emph{consistent} condition, both cues agree with true depth. In the \emph{conflict} condition, the designated cue favors the opposite depth ordering. In the \emph{decorrelated} condition, we balance cue directions across depth labels, so that the closer object is equally likely to appear larger or smaller, and higher or lower, than the farther object.

VLMs' behaviors are quantified with \emph{depth accuracy}, which is the fraction of examples where the model answers the truly closer object, with 50\% corresponding to random guessing.

\begin{figure}[h]
\centering
\includegraphics[width=\linewidth]{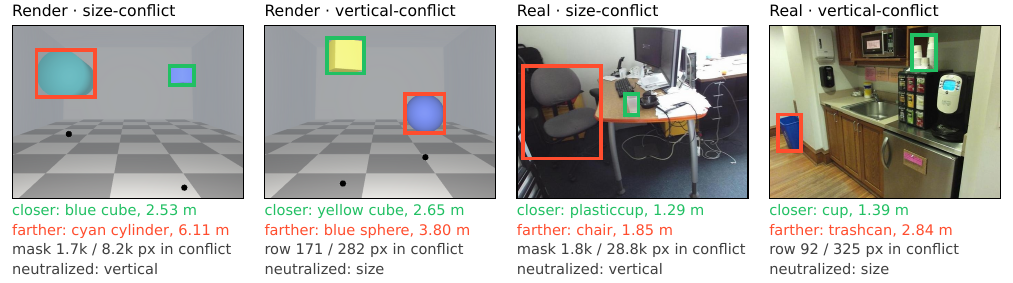}
\vspace{-15pt}
\caption{\textbf{Examples in which a 2D cue conflicts with true depth.} The black dot marks each object's position projected straight down onto the floor. Bounding boxes highlight the queried objects.}
\label{fig:stimuli}
\end{figure}

We first create rendered scenes where depth, object size, and vertical position can be set independently. We manipulate the physical size and image-plane position of the two objects while preserving their depth ordering, allowing each cue to agree with or oppose depth. Since object shape and perspective projection can make scene parameters differ from their final appearance, we measure projected size and vertical position directly from the rendered pixels and adjust each scene until the intended relation is satisfied.

We also construct a real-world-image setting using photographs and sensor depth from SUN RGB-D~\citep{song2015sun}. Sensor depth is used as a pseudo-label to determine which object is truly closer, while projected size and vertical position are measured directly from the image. For each object pair, we compare the ordering implied by each 2D cue with the sensor-depth ordering. A pair is consistent when the cue favors the truly closer object and conflict when it favors the farther one. This yields an analogous depth--cue comparison in natural photographs, although other scene properties are not independently controlled.

\subsection{VLMs Follow 2D Cues When They Conflict with Depth}

\begin{figure}[h]
\centering
\centering
\includegraphics[width=0.9\linewidth]{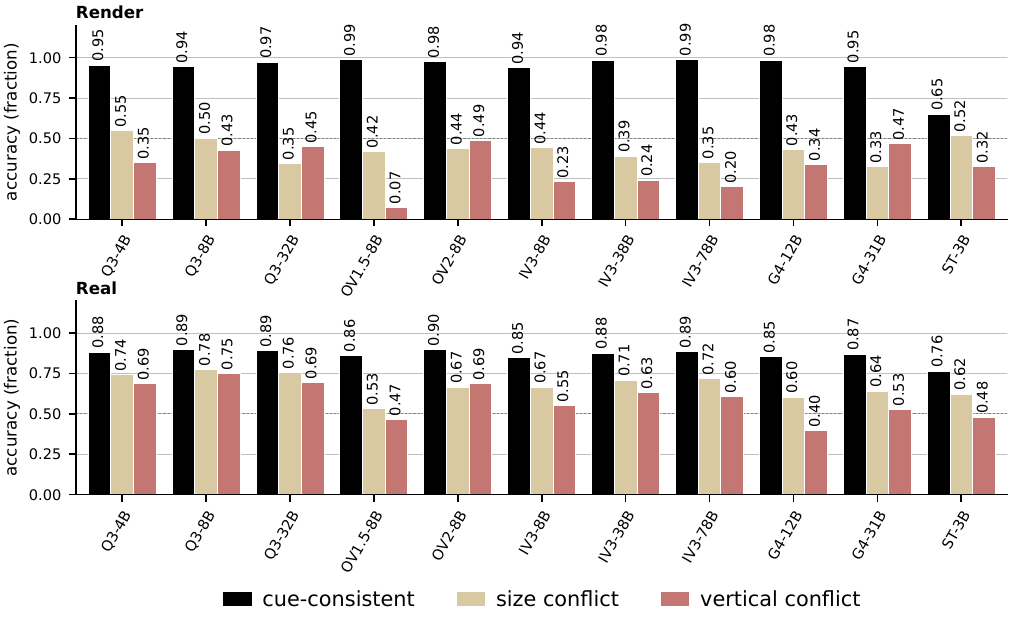}
\vspace{-10pt}
\caption{Depth accuracy drops across model families when image-plane cues
conflict with true depth.}
\label{fig:gapmatrix}
\end{figure}

A consistent pattern emerges across all five model families. Depth accuracy drops sharply when either 2D cue conflicts with true depth (Figure~\ref{fig:gapmatrix}). On rendered scenes, \qeight{} falls to 42.8\% under vertical conflict, and LLaVA-OneVision-1.5-8B to 7.4\%, both below chance. Spatial fine-tuning does not remove the effect. SpaceThinker, a Qwen2.5-VL-3B checkpoint trained on synthetic spatial reasoning traces~\citep{spacethinker}, reaches only 51.8\% and 32.4\% under the two conflict conditions.

Vertical position is generally the stronger cue. Its conflict produces a larger accuracy drop than size conflict in eight of eleven models, spanning all five families. The same pattern appears in natural photographs, though with smaller gaps because naturally occurring conflicts are less extreme than those in the rendered setting. Together, these results show that when 2D cues disagree with true depth, model predictions are systematically pulled toward the cue-supported answer.

\begin{figure}[h]
\centering
\begin{subfigure}{0.33\linewidth}
    \centering
    \includegraphics[width=\linewidth]{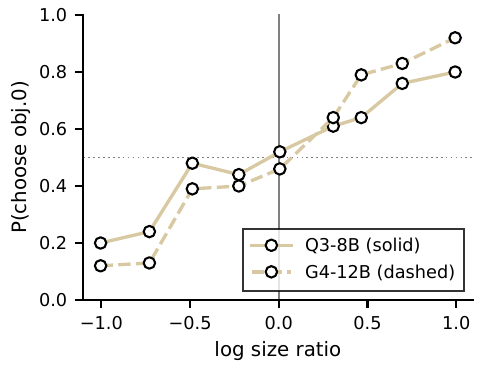}
    \caption{Size cue.}
\end{subfigure}\hfill
\begin{subfigure}{0.33\linewidth}
    \centering
    \includegraphics[width=\linewidth]{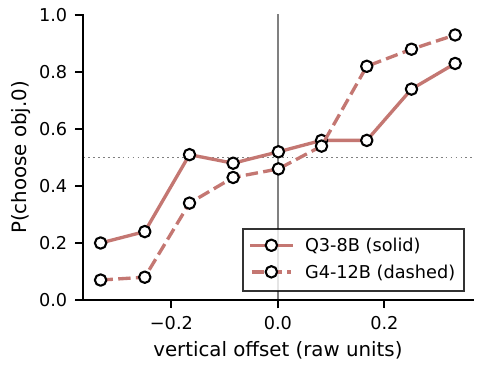}
    \caption{Vertical cue.}
\end{subfigure}\hfill
\begin{subfigure}{0.33\linewidth}
    \centering
    \includegraphics[width=\linewidth]{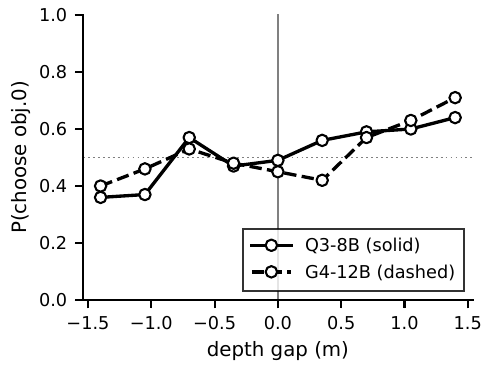}
    \caption{True depth.}
\end{subfigure}
\vspace{-5pt}

\caption{Impact of increasing the object's size, adjusting its vertical position, or changing its depth on the model's probability of selecting it as the closest.}
\label{fig:psych}
\end{figure}

Cue influence changes gradually with cue strength in rendered scenes (Figure~\ref{fig:psych}).
We hold the objects' true depths fixed and vary their apparent size or vertical position. For example, suppose object A is farther away than object B. As A is made to appear larger, the model becomes more likely to choose A as closer, although the correct answer remains B. Lowering A in the image produces a similar shift. These rendered-scene curves show that stronger misleading cues progressively pull decisions away from the correct depth order. Since size and vertical position are measured in different units, we use the curves to establish graded cue dependence rather than compare their raw slopes.

Separately, we examine how often these cues agree with depth in the real-world image pool before selecting conflict cases. Among object pairs whose vertical separation exceeds our specified threshold, the lower object is closer in 76.3\% of cases. Among pairs whose apparent-size difference exceeds the corresponding threshold, the larger object is closer in 58.1\% of cases. This ordering is consistent with the stronger behavioral influence of vertical position, although it does not establish its cause.

The pattern also appears across model scales and on external
spatial evaluations
(Appendices~\ref{app:external} and~\ref{app:benchmark}).
We next ask whether the correct depth relation remains
represented despite these cue-driven errors.

%% file: sections_0907/probing.tex
\section{Representation of Depth Order}
\label{sec:present}


\subsection{Linear Readout of the Depth Trace}

We read depth order from the representations of the two queried objects.
Visual tokens within each segmentation mask are mean-pooled into object
representations $c_A$ and $c_B$, and we define their contrast as
\begin{equation}
d = c_A - c_B.
\label{eq:object_contrast}
\end{equation}

At each layer, we fit a linear probe~\citep{probes} to object contrasts $d_i$ and their depth-order labels $y_i\in\{0,1\}$, where $y_i=1$ means that object $A$ is closer than object $B$. We standardize each contrast as $\tilde d_i=(d_i-\mu)/s$, where $\mu$ and $s$ are the per-dimension means and scales computed on the calibration training fold; division is elementwise.

The probe uses logistic regression to predict depth order:
\begin{equation}
p(y_i=1\mid d_i;w,b)=\sigma(w^\top\tilde d_i+b),
\label{eq:depth_probability}
\end{equation}
where $w$ is the weight vector, $b$ is the intercept, and
$\sigma$ is the sigmoid function. We fit these parameters by
maximizing the regularized log-likelihood:
\begin{equation}
(w^*,b^*)
=
\arg\max_{w,b}
\left[
\frac{1}{N}\sum_{i=1}^{N}
\log p(y_i\mid d_i;w,b)
-\frac{\lambda}{2}\|w\|_2^2
\right].
\label{eq:depth_probe}
\end{equation}
The likelihood term rewards correct depth-order predictions, while the regularization term limits the magnitude of the weights. The fitted weight $w^*$ is normal to the classifier's decision boundary in standardized feature space.

We map this direction back to the original hidden-state
coordinates and unit-normalize it:
\begin{equation}
v=\frac{w^*/s}{\|w^*/s\|_2}.
\label{eq:trace_direction}
\end{equation}
For a new object pair, we measure its contrast along $v$:
\begin{equation}
z=v^\top d,
\qquad d=c_A-c_B.
\label{eq:depth_readout}
\end{equation}
A positive $z$ favors object $A$ as closer, while a negative
$z$ favors object $B$. We call $z$ the \emph{depth trace}
and $v$ the \emph{trace direction}.
\begin{figure}[h]
    \centering
    \includegraphics[width=\linewidth]{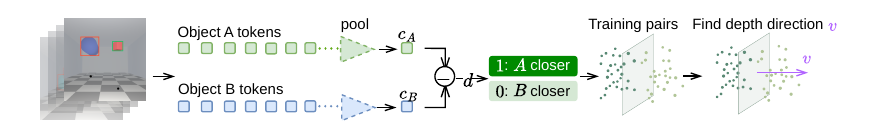}
    \vspace{-10pt}
    \caption{\textbf{Reading depth from object representations.}
    Object tokens are pooled and contrasted to fit the depth
    direction $v$. Point clouds and separating planes are schematic.
    }
    \label{fig:probe-fitting}
\end{figure}
\vspace{-5pt}
The fitted classifier uses the standardized features and intercept in Eq.~\ref{eq:depth_probability}. For the intervention, which strengthens the depth signal by updating the two objects' visual representations, we instead use the offset-free readout $z$ in Eq.~\ref{eq:depth_readout}, retaining the direction without the classifier's centering and intercept offsets. Classifier accuracy and intervention sign-readout accuracy are therefore evaluated separately.

Probes are fitted on the decorrelated condition, where apparent size and vertical position do not individually predict depth order. Full fitting details, including the regularization strength and standardization procedure, are provided in Appendix~\ref{app:protocols}.

The object-level readout is highly accurate. On Qwen3-VL-8B, probe accuracy remains above 97.3\% at every language layer above the embeddings on rendered scenes and ranges from
82.9\% to 91.7\% on real-world photographs. We test whether this readout reflects visual depth information rather than the two target cues or object-category priors, such as expecting a cup to be closer than a sofa regardless
of the image. On the decorrelated calibration split, neither apparent size nor vertical position alone predicts depth order. 
With the depth direction, question, and object-token positions held fixed, replacing the image sharply reduces depth-readout accuracy (Table~\ref{tab:image-controls}). On real-image vertical conflicts, Qwen3-VL-8B falls from 82.6\% with the original image to 40.3\% with a grey image. Mismatched-image controls show a similar reduction. The readout also generalizes to unseen category combinations. On the real-world image held-out fold, Qwen3-VL-8B achieves 84.4\% depth sign accuracy on pairs whose category combination is absent from the probe's training fold, compared with 84.5\% overall. These results show that the readout depends on the current image, even when apparent size or vertical position favors the wrong answer.

Where depth becomes linearly accessible differs across architectures. On \qeight{}, the pre-projection probe reaches 74.0\%, and accuracy is already 99.7\% at the language-model input embeddings and remains near ceiling thereafter. On \gtwelve{}, the corresponding readouts are 62.5\% and 66.0\%, with near-ceiling accuracy emerging after the first decoder block. These probes locate where depth becomes linearly accessible, rather than where depth information first exists.

\subsection{Depth Information at Decision States}

\begin{figure}[h]
\centering
\includegraphics[width=0.9\linewidth]{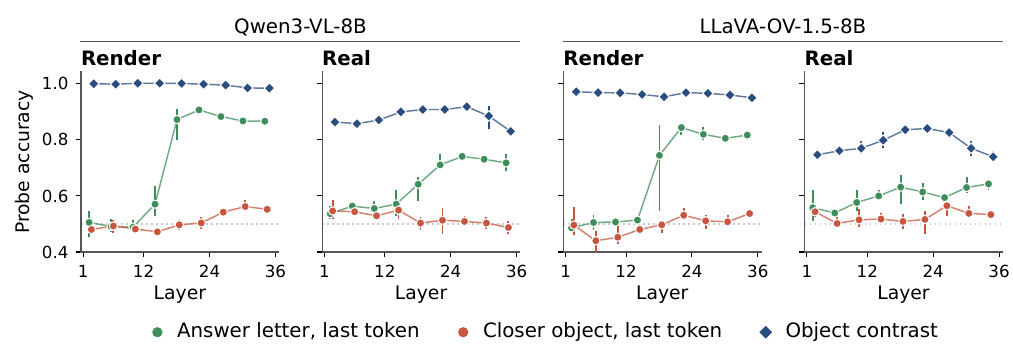}
\vspace{-10pt}
\caption{\textbf{Depth readouts from answer states and object representations.} Points average four consecutive layers; bars indicate within-bin minima and maxima, not confidence intervals.}
\label{fig:probes}
\end{figure}

We next ask whether depth information is accessible where the model prepares its answer. We call the hidden state at the last input-token position the \emph{answer state}, since the model uses this position to predict the next token. We probe this state at each layer and compare it with the two objects' visual representations (Figure~\ref{fig:probes}).

We train two separate probes on the same answer states. The \emph{closer-object probe} predicts which of the two queried objects is truly closer. The \emph{answer-letter probe} predicts the letter assigned to that object in the answer options, regardless of whether the model itself answers correctly. For example, if the mug is closer than the sofa, the closer-object target remains the mug when the options are swapped from ``A: mug, B: sofa'' to ``A: sofa, B: mug'', whereas the answer-letter target changes from ``A'' to ``B''. This comparison tests whether the correct depth order is more accessible as an object relation or as an answer choice.

On Render, probes struggle to identify the closer object from the answer state, with accuracy generally near chance. Yet probes can predict the correct answer letter from the same states: accuracy rises sharply around middle layers and peaks at 91.3\% in Qwen and 85.3\% in OneVision. The same distinction appears on Real, although answer-letter accuracy is lower, peaking at 75.3\% and 67.2\%, respectively. Thus, the answer state can support accurate decoding of the correct option even when the closer object's identity is difficult to decode directly.

Depth order is more consistently accessible from object-representation differences in both models. Decoding remains high throughout the stack on Render, while on Real it peaks in the middle layers and declines toward the final layers.

Decodability does not guarantee a correct answer. In a separate evaluation on rendered vertical-conflict examples, Qwen's object-level depth readout is correct on 99.0\% of examples, while the model answers correctly on only 42.8\%. This gap motivates testing whether strengthening depth information in object representations can change the final decision.

%% file: sections_0907/causal.tex
\section{Testing the Influence of Depth Information}
\label{sec:causal}

The previous section shows that the correct depth relation can remain strongly represented even when the final answer follows a conflicting 2D cue. Probing alone, however, does not show whether this representation influences the decision. We therefore intervene directly on the fitted depth direction while keeping the input and model weights fixed.

\subsection{Write-back intervention to strengthen the depth trace}
\label{sec:intervention}

We use the depth direction identified in Section~\ref{sec:present}. On rendered scenes, the direction is fitted on the decorrelated calibration split, where apparent size and vertical position do not predict depth order. For real-world photographs, we refit the direction on the corresponding development subset split using DepthAnything~3~\citep{lin2025depth} to provide pseudo-depth orders. The evaluation example's depth label is never used to orient the intervention.
\vspace{-15pt}
\begin{figure}[h]
\centering
\includegraphics[width=0.9\linewidth]{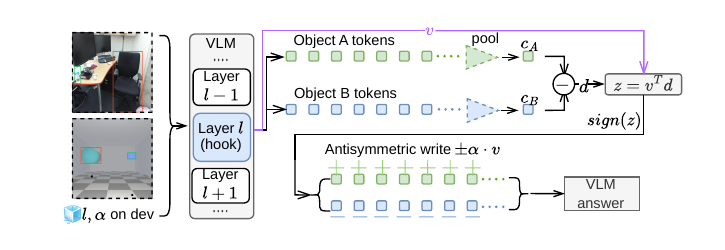}
\vspace{-10pt}
\caption{\textbf{Intervention to strengthen the depth trace.}
The sign of the pre-intervention depth trace $z$ determines
opposite updates to the two queried objects along $v$.
The intervention uses no ground-truth depth label.}
\label{fig:intervention-method}
\end{figure}

\vspace{-5pt}
We select the intervention layer $\ell$ and strength $\alpha$ on development data. For each candidate setting, we measure
the fractions of the size and vertical conflict gaps recovered,
$R_{\mathrm{size}}(\ell,\alpha)$ and
$R_{\mathrm{vertical}}(\ell,\alpha)$.
We score the setting by the weaker recovery:
\begin{equation}
R_{\mathrm{balanced}}(\ell,\alpha)
=
\min\left\{
R_{\mathrm{size}}(\ell,\alpha),
R_{\mathrm{vertical}}(\ell,\alpha)
\right\}.
\label{eq:balanced_recovery}
\end{equation}
Among settings satisfying the development-set constraint on
cue-consistent accuracy, we select the one with the largest
balanced recovery. This favors settings that improve both
conflict conditions. We then freeze $\ell$ and $\alpha$
before evaluation. The search ranges and accuracy constraint
are specified in Appendix~\ref{app:protocols}.

At layer $\ell$, we mean-pool the visual tokens of the two
queried objects into $c_A$ and $c_B$, form their contrast
$d=c_A-c_B$, and compute $z=v^\top d$.
As illustrated in Figure~\ref{fig:intervention-method}, we strengthen
this readout through opposite updates to the two objects.
Expressed in units of the layer's mean token norm, the updates are
\begin{equation}
\begin{aligned}
\text{Object $A$ tokens:}\quad
&+\alpha\,\operatorname{sign}(z)\,v,\\
\text{Object $B$ tokens:}\quad
&-\alpha\,\operatorname{sign}(z)\,v.
\end{aligned}
\label{eq:depth_writeback}
\end{equation}
Each update is multiplied by the mean token norm at that layer
before being added to the corresponding token states.
When $z>0$, the edit moves object $A$ along $v$ and object $B$
against it; when $z<0$, both updates reverse.
The intervention therefore strengthens the ordering read
from the current image rather than supplying a ground-truth
ordering.
Only the queried object tokens are directly modified at the
selected layer. Subsequent layers process the edited states
normally, while the input and VLM parameters remain fixed.

\subsection{Strengthening the Depth Trace Changes the Decision}
\label{sec:causal-effect}

\begin{table}[h]
\centering
\footnotesize
\setlength{\tabcolsep}{3pt}
\caption{\textbf{Effect of depth-trace intervention across models.}
Accuracy (\%) before and after intervention on rendered and natural
images. Each setup is selected on held-out development data and
frozen before evaluation. (`van.' is vanilla and `int.' is the intervened model)}
\label{tab:permodel}
\begin{tabular}{l cc cc cc cc cc cc}
\toprule
 & \multicolumn{6}{c}{Rendered}
 & \multicolumn{6}{c}{Real} \\
\cmidrule(lr){2-7}\cmidrule(lr){8-13}
 & \multicolumn{2}{c}{conflict-size}
 & \multicolumn{2}{c}{conflict-vert.}
 & \multicolumn{2}{c}{consistent}
 & \multicolumn{2}{c}{conflict-size}
 & \multicolumn{2}{c}{conflict-vert.}
 & \multicolumn{2}{c}{consistent} \\
 & van. & int. & van. & int. & van. & int.
 & van. & int. & van. & int. & van. & int. \\
\midrule
Qwen3-VL-8B
 & \pid{1863} & \pid{2162} & \pid{1864} & \pid{2163} & \pid{1862} & \pid{2161}
 & \pid{155} & \pid{452} & \pid{156} & \pid{453} & \pid{154} & \pid{454} \\
Qwen3-VL-4B
 & \pid{1859} & \pid{2158} & \pid{1860} & \pid{2159} & \pid{1858} & \pid{2157}
 & \pid{151} & \pid{516} & \pid{152} & \pid{517} & \pid{150} & \pid{518} \\
Qwen3-VL-32B
 & \pid{1867} & \pid{2166} & \pid{1868} & \pid{2167} & \pid{1866} & \pid{2165}
 & \pid{159} & \pid{463} & \pid{160} & \pid{464} & \pid{158} & \pid{465} \\
LLaVA-OV-1.5-8B
 & \pid{1871} & \pid{2170} & \pid{1872} & \pid{2171} & \pid{1870} & \pid{2169}
 & \pid{171} & \pid{481} & \pid{172} & \pid{482} & \pid{170} & \pid{483} \\
LLaVA-OV-2-8B
 & \pid{1875} & \pid{2174} & \pid{1876} & \pid{2175} & \pid{1874} & \pid{2173}
 & \pid{175} & \pid{487} & \pid{176} & \pid{488} & \pid{174} & \pid{489} \\
\midrule
InternVL3-8B
 & \pid{1879} & \pid{2178} & \pid{1880} & \pid{2179} & \pid{1878} & \pid{2177}
 & \pid{1181} & \pid{1338} & \pid{1182} & \pid{1339} & \pid{1180} & \pid{1340} \\
InternVL3-38B
 & \pid{1883} & \pid{1931} & \pid{1884} & \pid{1932} & \pid{1882} & \pid{1933}
 & \pid{1189} & \pid{1341} & \pid{1190} & \pid{1342} & \pid{1188} & \pid{1343} \\
Gemma-4-12B
 & \pid{1887} & \pid{2186} & \pid{1888} & \pid{2187} & \pid{1886} & \pid{2185}
 & \pid{163} & \pid{469} & \pid{164} & \pid{470} & \pid{162} & \pid{471} \\
Gemma-4-31B
 & \pid{1891} & \pid{1937} & \pid{1892} & \pid{1938} & \pid{1890} & \pid{1939}
 & \pid{167} & \pid{475} & \pid{168} & \pid{476} & \pid{166} & \pid{477} \\
\bottomrule
\end{tabular}
\end{table}

Strengthening the depth trace moves conflict decisions toward the depth order already represented by the model. On rendered scenes, for \qeight{}, accuracy rises from \pidpct{1863} to \pidpct{2162} under size conflict and from \pidpct{1864} to \pidpct{2163} under vertical conflict, while cue-consistent accuracy remains high (\pidpct{1862} to \pidpct{2161}). 
The update shifts the model's answer toward the object identified as closer by its pre-intervention depth readout.

The effect extends beyond \qeight{} but varies substantially across architectures (Table~\ref{tab:permodel}). Qwen3-VL and LLaVA-OneVision show large shifts under conflict, as does InternVL3-8B, whereas the effect is weaker for InternVL3-38B and Gemma-4. A linearly readable depth trace therefore does not imply an equally strong causal handle in every model. 

\begin{figure}[h]
    \centering
    \includegraphics[width=\linewidth]{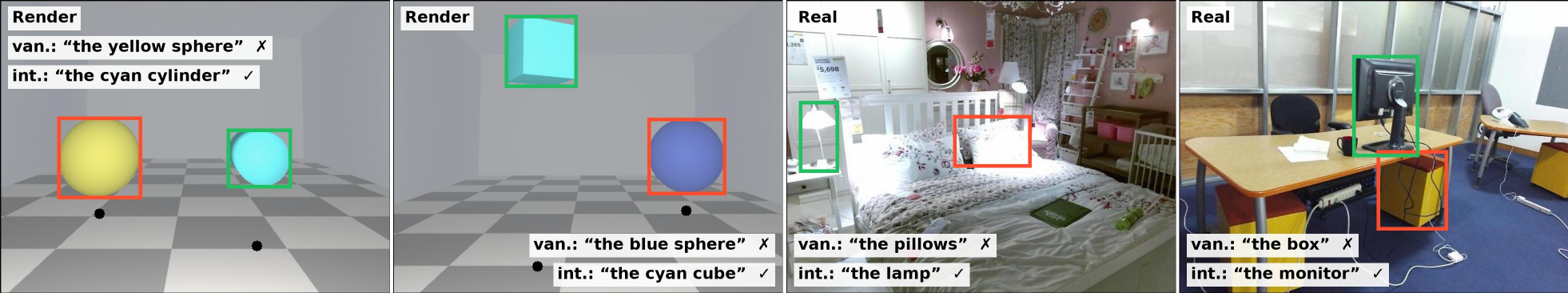}
    \caption{
    Example corrections on rendered scenes and natural photographs.
    }
    \label{fig:casepair}
\end{figure}

On real photographs, write-back increases observed accuracy in both conflict conditions for Qwen3-VL and LLaVA-OneVision, with larger gains in some models than others. The effect is mixed or negative for InternVL3 and Gemma-4, and cue-consistent accuracy often decreases. These results indicate that the benefit depends on the model and evaluation setting. The 95\% confidence intervals for accuracy changes are reported in Tables~\ref{tab:render-ci} and~\ref{tab:real-ci}.
Figure~\ref{fig:casepair} illustrates conflict cases in which
write-back changes a cue-supported error into the correct
depth judgment.

\begin{figure}[h]
    \centering
    \includegraphics[width=\linewidth]{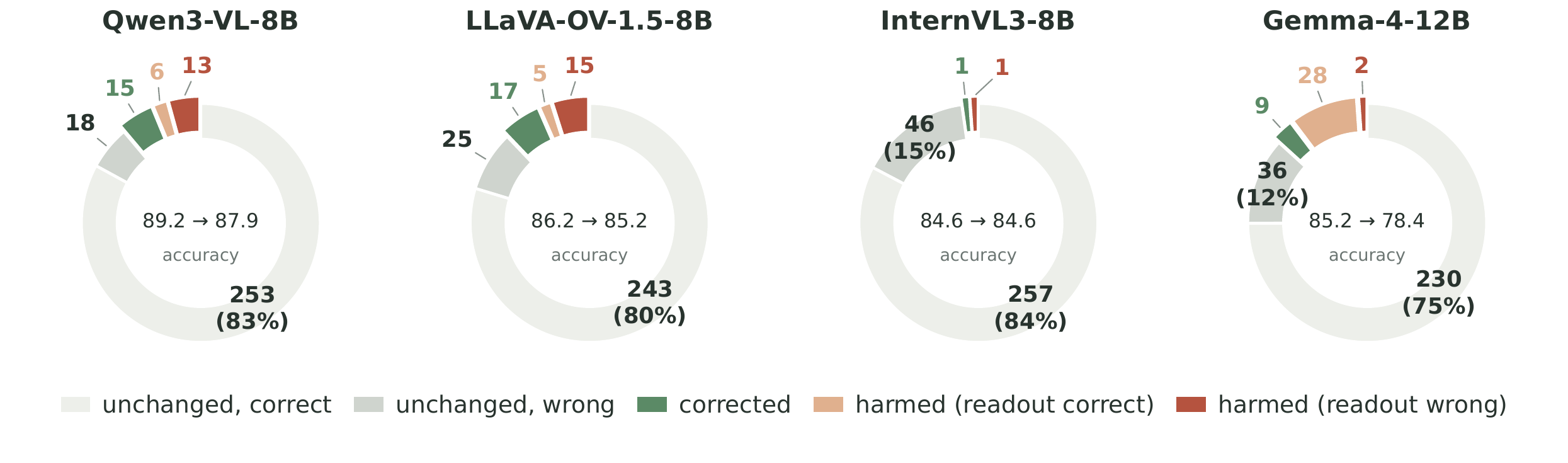}
    \vspace{-20pt}
    \caption{\textbf{How answers change after the write-back intervention on consistent real-world images.} Changes from correct to incorrect are categorized by whether the pre-intervention readout identified the closer object correctly. Numbers show the counts of corresponding images.}
    \label{fig:consistent-readout-errors}
\end{figure}

The source of consistent-split degradation varies across models (Figure~\ref{fig:consistent-readout-errors}). For Qwen3-VL-8B and LLaVA-OneVision-1.5-8B, most newly introduced errors occur when the pre-intervention depth readout identifies the farther object as closer, orienting the write-back in the wrong direction. In Gemma-4-12B, however, most new errors occur even when the readout identifies the closer object correctly. Thus, an incorrect update sign alone cannot explain the degradation across architectures. InternVL3-8B changes few answers and shows no net accuracy decline.

\subsection{Ablations}
\label{sec:causal-controls}


We test the intervention on Qwen3-VL-8B in both domains using
1) cue-avoidant prompting (instructing the model not to use cues), 2) predicted depth input (providing external depth map to the model), 3) random directions of the $v$ vector,
and alternative write locations (answer token vs. object contrast) (Table~\ref{tab:causal-controls}).
Neither explicitly discouraging the two cues nor supplying a
predicted depth map reproduces the gains from object-level write-back.

The effect is also specific to the internal representation being manipulated. Replacing the depth direction with random directions leaves conflict performance near the unmodified model. Writing the same depth direction at the answer token rather than antisymmetrically at the two queried objects likewise produces little selective improvement. Reversing the orientation of the depth write-back produces the opposite behavioral shift. 

\begin{table}[h]
\centering
\footnotesize
\setlength{\tabcolsep}{3.2pt}
\caption{Ablation study on Qwen3-VL-8B. Values are answer accuracy (\%). The write-back intervention gives the largest improvement for size and vertical conflicts.}
\label{tab:causal-controls}
\begin{tabular}{l ccc ccc}
\toprule
 & \multicolumn{3}{c}{Rendered}
 & \multicolumn{3}{c}{Real} \\
\cmidrule(lr){2-4}\cmidrule(lr){5-7}
 & size & vertical & consistent
 & size & vertical & consistent \\
\midrule
Vanilla
 & \pid{1863} & \pid{1864} & \pid{1862}
 & \pid{155} & \pid{156} & \pid{154} \\

Cue-avoidant CoT
 & \pid{2014} & \pid{2015} & \pid{2212}
 & \pid{1249} & \pid{1250} & \pid{1251} \\

Depth input
 & \pid{2022} & \pid{2023} & \pid{2216}
 & \pid{557} & \pid{558} & \pid{556} \\

Random direction
 & \pid{2273} & \pid{2274} & \pid{2275}
 & \pid{2352} & \pid{2353} & \pid{2354} \\

Answer-token write
 & \pid{2282} & \pid{2283} & \pid{2284}
 & \pid{2361} & \pid{2362} & \textbf{\pid{2363}} \\

\midrule
Write-back intervention
 & \textbf{\pid{2162}} & \textbf{\pid{2163}} & \textbf{\pid{2161}}
 & \textbf{\pid{2343}} & \textbf{\pid{2344}} & \pid{454} \\
\bottomrule
\end{tabular}
\end{table}


\vspace{-10pt}
\begin{figure}[h]
\centering
\includegraphics[width=0.9\linewidth]{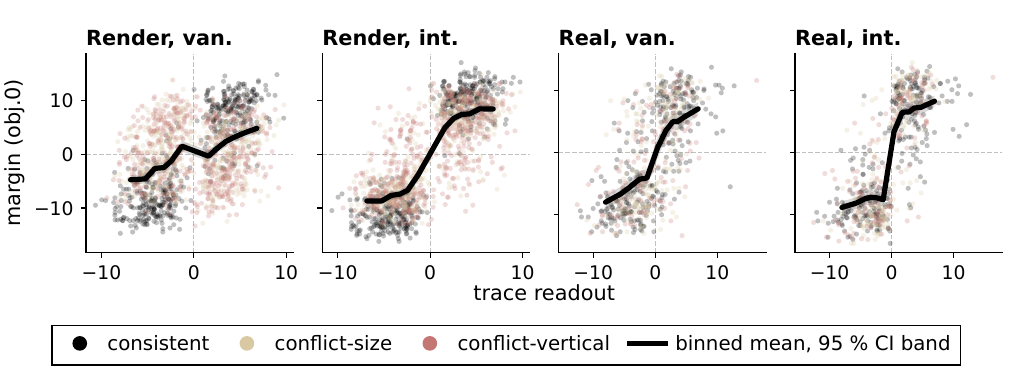}
\caption{Depth readout versus answer-logit margin before and after write-back. The margin is the logit for the option naming object 0 minus that for naming object 1; positive values favor object 0. Black curves show binned means with 95\% confidence bands.}
\label{fig:trace-readout}
\end{figure}

Figure~\ref{fig:trace-readout} shows how write-back changes the relationship between the internal depth readout and the final answer. On Render, the vanilla answer margin varies only weakly with the trace readout, whereas write-back produces a clearer separation between negative and positive readouts. On Real, the vanilla predictions already track the readout, and write-back sharpens this transition around $z=0$. 
Together, these results show that object-level write-back shifts
the model's answer toward the depth ordering decoded before intervention.

%% file: sections_0907/limitations.tex
\section{Limitations}
\label{sec:limitations}

Our analysis focuses on pairwise depth judgments and two common image-plane
cues, apparent size and vertical position. Real-world images contain additional
depth-relevant signals that cannot be isolated as cleanly as in controlled
rendering, which may contribute to the smaller effects observed there.
The causal effect also varies across model families and domains. Extending
the analysis to additional depth cues and more general spatial reasoning
is an important direction for future work.

%% file: backmatter/ai_statement.tex
\section{Statement on the use of AI assistance}
\label{app:ai}
The experimental program and manuscript preparation used a coding agent
under the authors' direction. The agent wrote and ran scripts and
organized result artifacts. Experimental design and decisions were the
authors'. The authors take responsibility for the manuscript and results.

%% file: backmatter/repro_statement.tex
\section{Reproducibility Statement}
\label{app:repro}
Data construction and answer scoring are described in
Appendix~\ref{app:benchmark}; probe fitting, object-level write-back,
and controls are detailed in Appendix~\ref{app:protocols}.
Dataset examples appear in Appendix~\ref{app:stimuli}.
The working repository contains the experimental code, result artifacts,
checkpoint revisions, and image-level manifests recording object pairs,
option assignments, depth labels, and measured cue values.
The main-paper intervention results use object-only write-back,
with directions and intervention settings fixed before evaluation.

%% file: appendix/c_external.tex
\section{External benchmarks}
\label{app:external}

\subsection{External diagnosis}
SpatialRGPT-Bench~\citep{spatialrgpt} and SpatialTunnel reproduce the
cue-conflict pattern on data we did not construct. On the 92
SpatialRGPT depth items, Qwen3-VL-8B scores \pidpct{590} on
cue-consistent items and \pidpct{592} on conflicts. The conflict subset
has only 13 items and mixed sources, so it is supporting evidence rather
than a standalone estimate. Five items are near-duplicates of our
evaluation images, including one calibration image.

\subsection{Intervention results}
Table~\ref{tab:external-results} applies the intervention after fitting
the direction on calibration images from the target benchmark. It helps
the small SpatialRGPT conflict subset and is nearly neutral on pooled
SpatialTunnel for Qwen3-VL-8B. It does not improve the other external
settings. In particular, accuracy falls on CV-Bench~\citep{cvbench} and on the
Qwen3-VL-32B SpatialTunnel pool. These negative results limit transfer
beyond the diagnostic distributions.

\begin{table}[htbp]
\centering
\small
\caption{External answer accuracy (\%). Pooled SpatialTunnel combines
976 phase and 1076 size items that support object masks. SpatialRGPT
conflict is directional and contains 13 items.}
\label{tab:external-results}
\begin{tabular}{llcc}
\toprule
Benchmark & Model & Vanilla & Intervention \\
\midrule
SpatialRGPT depth & Qwen3-VL-8B & \pid{644} & \pid{645} \\
SpatialRGPT conflict & Qwen3-VL-8B & \pid{592} & \pid{658} \\
SpatialTunnel pooled & Qwen3-VL-8B & \pid{2260} & \pid{2261} \\
SpatialTunnel pooled & Qwen3-VL-32B & \pid{2262} & \pid{2263} \\
CV-Bench depth & Qwen3-VL-8B & \pid{632} & \pid{633} \\
CV-Bench depth & Qwen3-VL-32B & \pid{638} & \pid{639} \\
\bottomrule
\end{tabular}
\end{table}

\begin{figure}[htbp]
\centering
\includegraphics[width=0.9\linewidth]{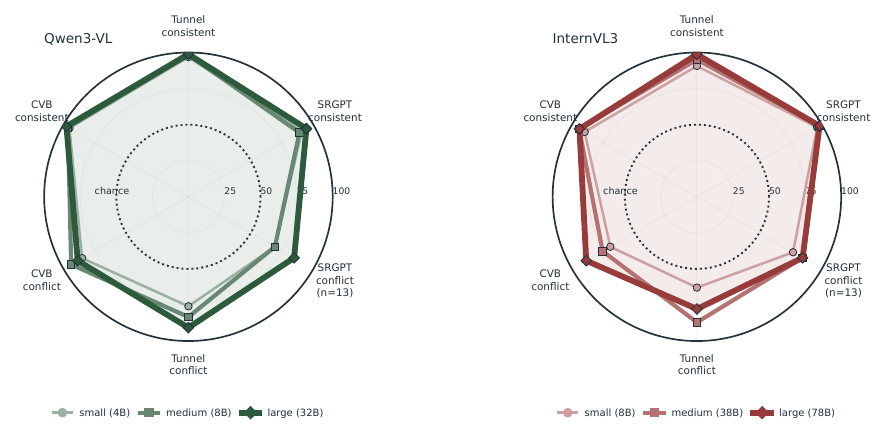}
\caption{Vanilla accuracy on three external benchmarks. The upper spokes
are cue-consistent and the lower spokes are cue-conflict conditions.}
\label{fig:radar}
\end{figure}

Figure~\ref{fig:radar} shows vanilla accuracy across model scales. The
SpatialRGPT conflict spoke has 13 items; the CV-Bench cue split was
constructed from measured image cues and the benchmark depth ordering.

%% file: appendix/h_benchmark_construction.tex
\section{Data and evaluation}
\label{app:benchmark}

\subsection{Rendered scenes}
Each scene contains two primitives in a room with a checkerboard floor.
Depth, angular size, and image position are sampled independently, while
physical size and world position are derived from them. The camera light,
depth fog, floor, and walls retain valid depth cues. Objects appear on
opposite sides of the image and do not occlude each other.

The dataset has 500 scenes in each of four splits. In the consistent
split, the closer object is larger and lower. In size conflict, it is
smaller and vertical position is neutralized. In vertical conflict, it
is higher and size is neutralized. In the decorrelated split, both cues
are independent of depth. A scene is rerendered up to four times until
the neutralized objects differ by at most 3\% in mask area or one pixel
in centroid row, as appropriate.

Images are $448\times448$. Depth ranges from 2.5 to 7.0\,m with at least
1.0\,m separation. Angular size ranges from 0.055 to 0.115 with at least
0.02 separation, and vertical position from $-0.18$ to $0.18$ with at
least 0.08 separation. Masks below 60 pixels are rejected. The base seed
is 0 and split $i$ uses seed $1000i$.

\subsection{Real photographs}
We rasterize SUN RGB-D polygons~\citep{song2015sun} and define object
depth as the median sensor depth inside each mask. A pair must differ by
at least 0.4\,m and a depth factor of 1.20. Box IoU is at most 5\%; each
mask covers at least 1500 pixels and at most one quarter of the image;
neither box is within five pixels of the border. Each category appears
only once in the image, and each image contributes at most three pairs.
The neutralized cue is sign-balanced within each conflict split.

\subsection{Evaluation}

We score each example by comparing the next-token logits for ``A'' and
``B'' at the final input position. Option order is randomized per item
and recorded in the manifest. Each input contains one user message with
the image and question, formatted using the model's chat template with
a generation prompt and no system prompt.

For rendered scenes, the question is:
``Two objects are visible in the image. Which object is closer to the
camera? A: the \{a\}. B: the \{b\}. Answer with only the letter A or B.''
For SUN RGB-D, it is:
``Which object is closer to the camera, the \{a\} or the \{b\}?
A: the \{a\}. B: the \{b\}. Answer with only the letter A or B.''
The harness tokenizes the bare answer strings without special tokens
and verifies that each is a single token. 

For SUN RGB-D, object depth is the median refined sensor depth within
the instance polygon; apparent size is mask area and vertical position
is centroid row. Cue-consistent pairs require the closer object to have
at least 1.25 times the area and appear lower by at least 7\% of image
height. Size-conflict pairs require its area ratio to be at most
$1/1.25$, with vertical separation at most 6\% of image height.
Vertical-conflict pairs require it to appear higher by at least 7\%,
with an area ratio in $[1/1.35,\,1.35]$. The decorrelated split balances
the four combinations of size and vertical cue directions.

The final splits contain 305 consistent pairs,
209 size-conflict pairs, 144 vertical-conflict pairs
, and 289 decorrelated pairs. The accompanying image-level manifest
records all 947 pairs, including scene identifiers, image and mask
paths, object identities, depth order, option assignment, and measured
cue values.

Patch shuffling permutes visual-token positions while retaining local
appearance information, including fog and shading. The grey-image
control retains the original object masks and therefore preserves the
positions used for token selection. Neither control removes every
possible source of spatial information.

\begin{figure}[htbp]
\centering
\begin{subfigure}[t]{0.47\linewidth}
  \includegraphics[width=\linewidth]{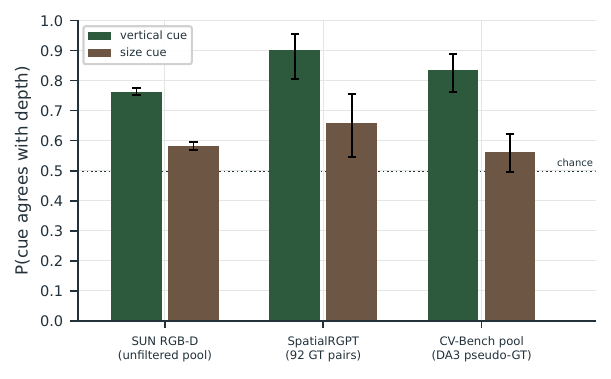}
  \caption{Agreement across datasets.}
\end{subfigure}\hfill
\begin{subfigure}[t]{0.47\linewidth}
  \includegraphics[width=\linewidth]{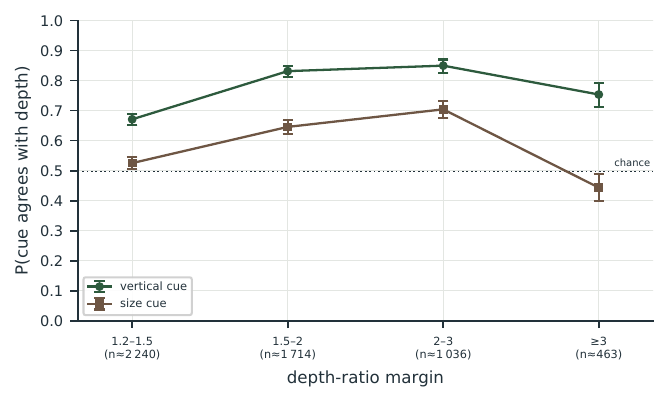}
  \caption{SUN RGB-D by depth ratio.}
\end{subfigure}
\caption{Agreement of image-plane cues with depth. Error bars are Wilson
95\% intervals; intervals in (b) use approximate reweighted sample sizes.}
\label{fig:cue-agreement}
\label{fig:baserates}
\label{fig:cue-margin-strata}
\end{figure}

Figure~\ref{fig:cue-agreement} reports cue--depth agreement and its
variation with depth ratio.

%% file: appendix/i_stimuli.tex
\subsection{Dataset examples}
\label{app:stimuli}
Each panel shows the first 16 recorded items in manifest order. Tile text
gives the item ID, answer assignment, depth order, mask area, and centroid
row. Rendered labels use scene depth; real labels use median sensor depth
inside each object mask.

\begin{figure}[h]
\centering
\begin{subfigure}[t]{0.48\linewidth}
  \centering
  \includegraphics[width=\linewidth,height=0.38\textheight,keepaspectratio]{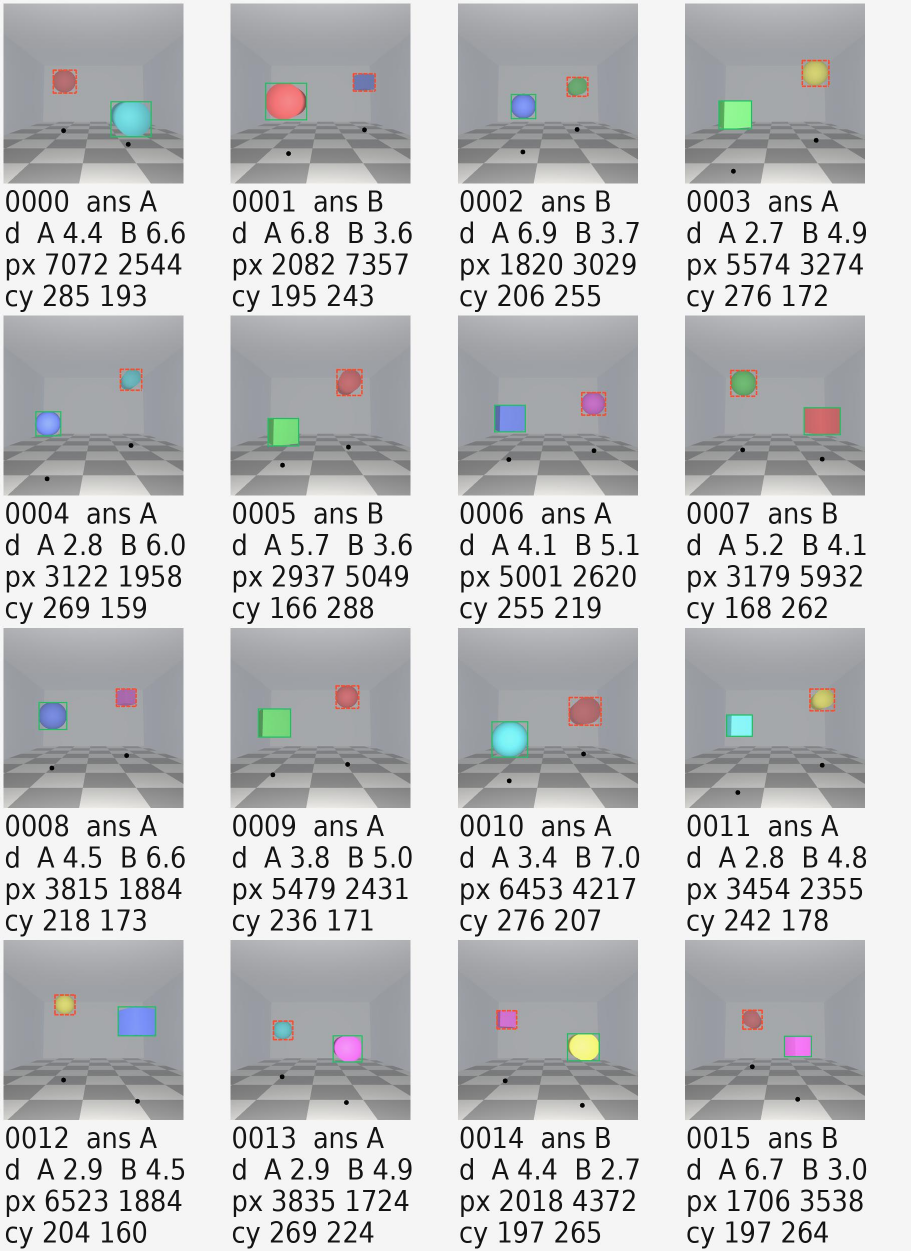}
  \caption{Consistent}
\end{subfigure}\hfill
\begin{subfigure}[t]{0.48\linewidth}
  \centering
  \includegraphics[width=\linewidth,height=0.38\textheight,keepaspectratio]{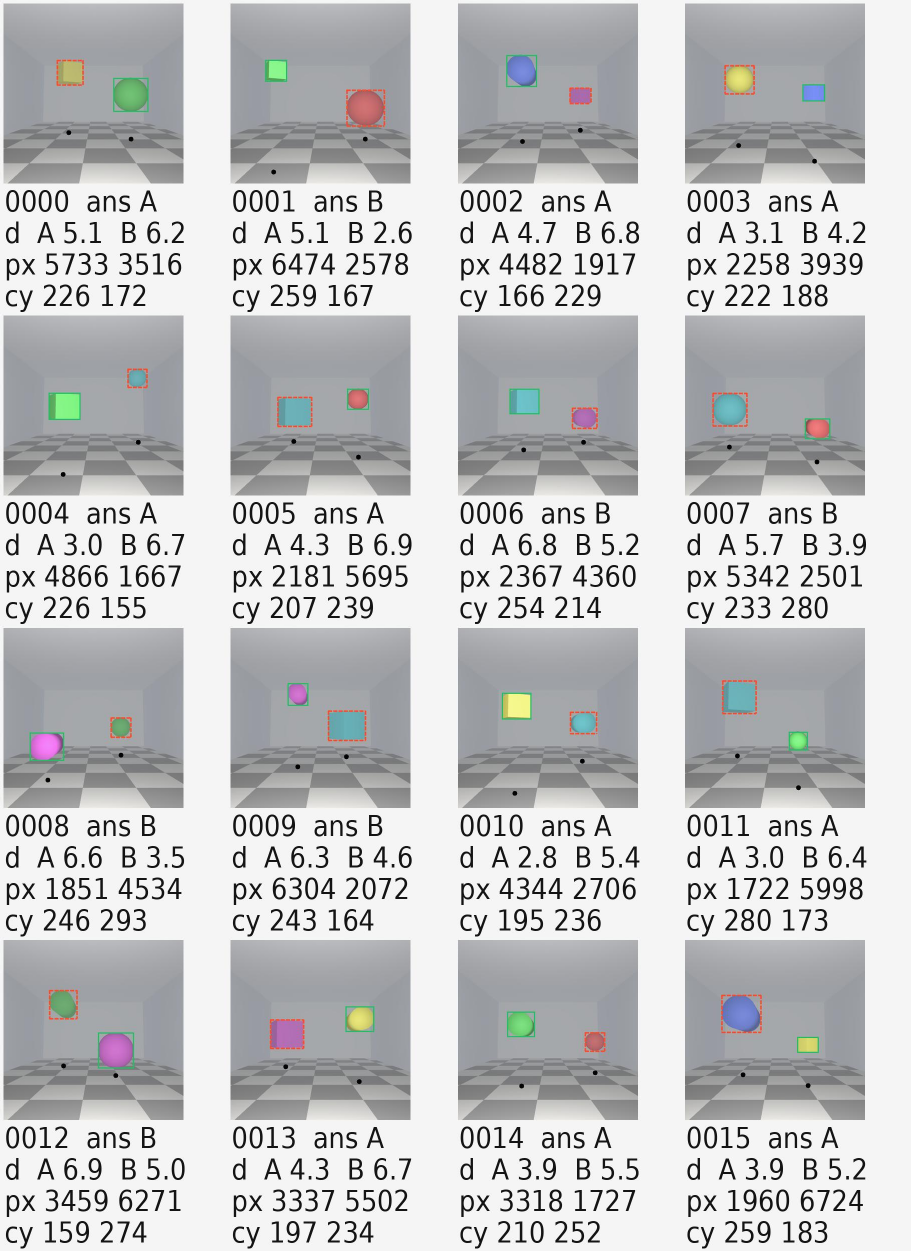}
  \caption{Decorrelated}
\end{subfigure}

\begin{subfigure}[t]{0.48\linewidth}
  \centering
  \includegraphics[width=\linewidth,height=0.38\textheight,keepaspectratio]{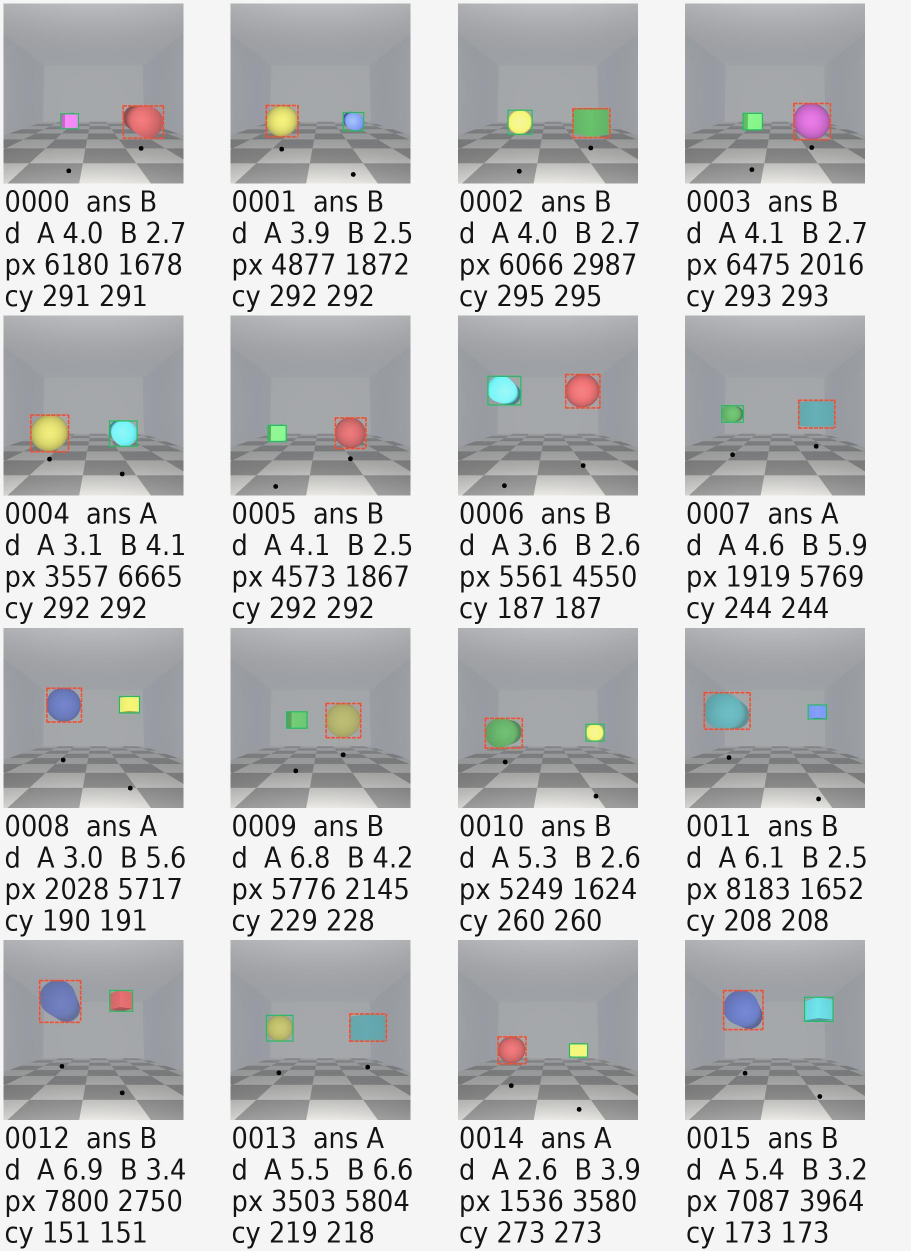}
  \caption{Size conflict}
\end{subfigure}\hfill
\begin{subfigure}[t]{0.48\linewidth}
  \centering
  \includegraphics[width=\linewidth,height=0.38\textheight,keepaspectratio]{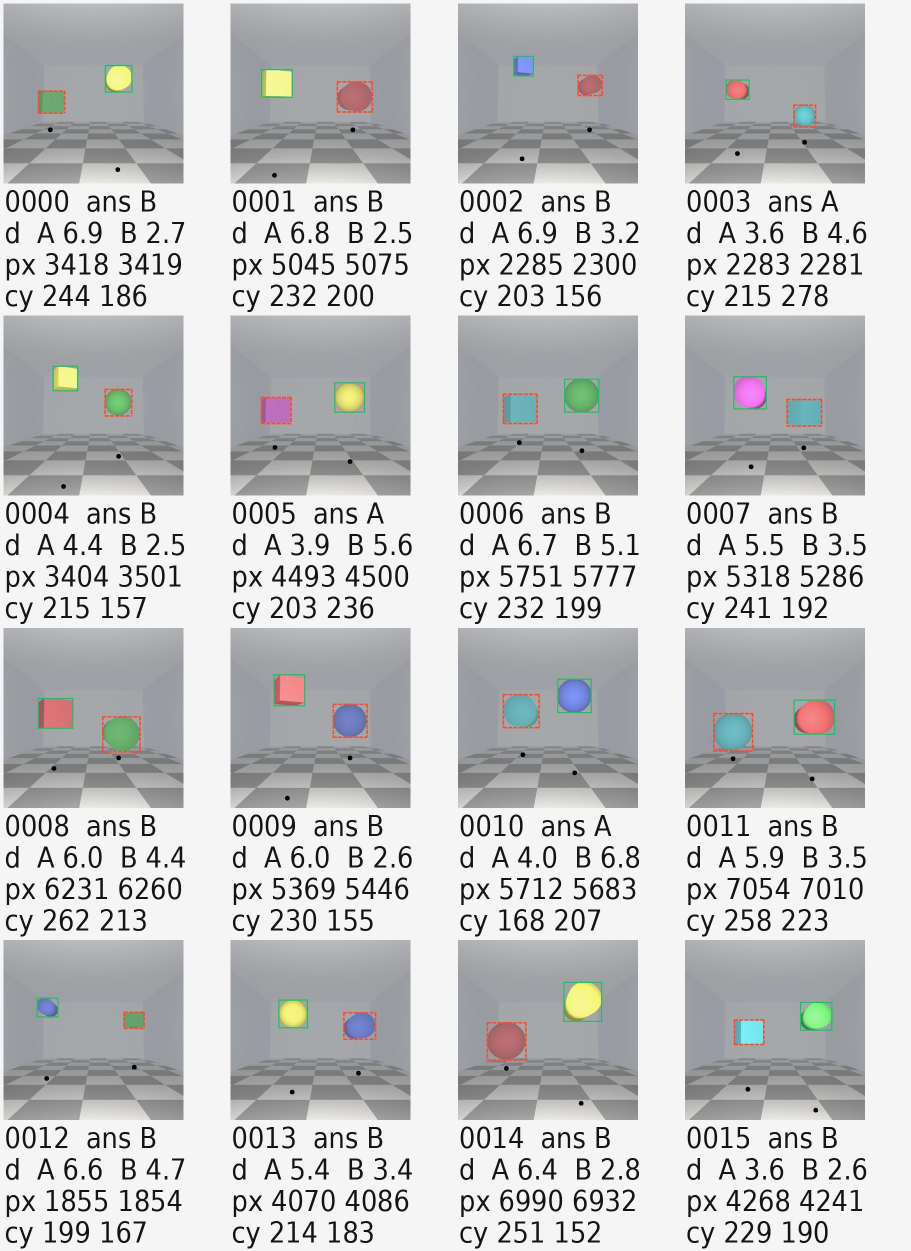}
  \caption{Vertical conflict}
\end{subfigure}
\caption{Rendered dataset examples. Depth, apparent size, and vertical
position are controlled independently.}
\label{fig:stimuli_rendered}
\end{figure}

\clearpage
\begin{figure}[h]
\centering
\begin{subfigure}[t]{0.48\linewidth}
  \centering
  \includegraphics[width=\linewidth,height=0.38\textheight,keepaspectratio]{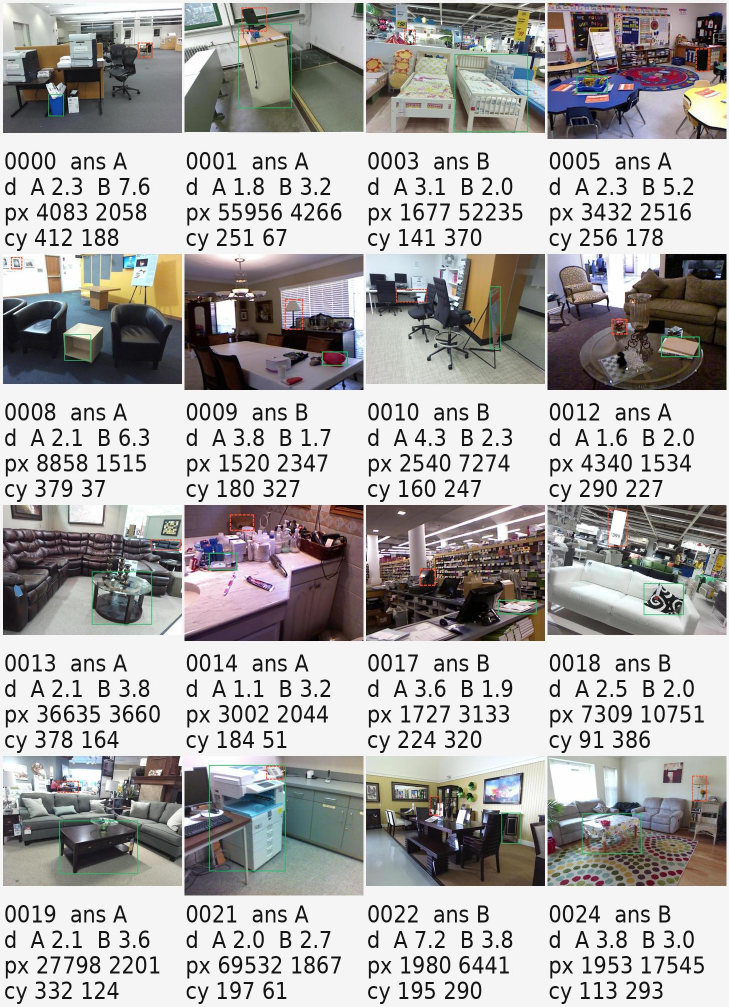}
  \caption{Consistent}
\end{subfigure}\hfill
\begin{subfigure}[t]{0.48\linewidth}
  \centering
  \includegraphics[width=\linewidth,height=0.38\textheight,keepaspectratio]{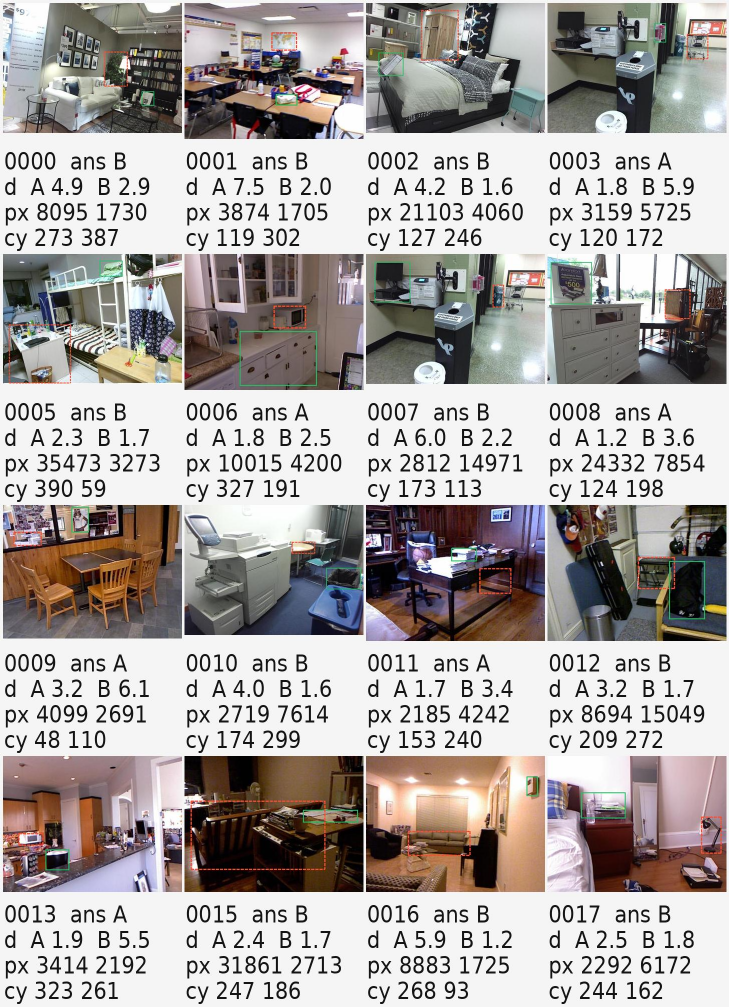}
  \caption{Decorrelated}
\end{subfigure}

\begin{subfigure}[t]{0.48\linewidth}
  \centering
  \includegraphics[width=\linewidth,height=0.38\textheight,keepaspectratio]{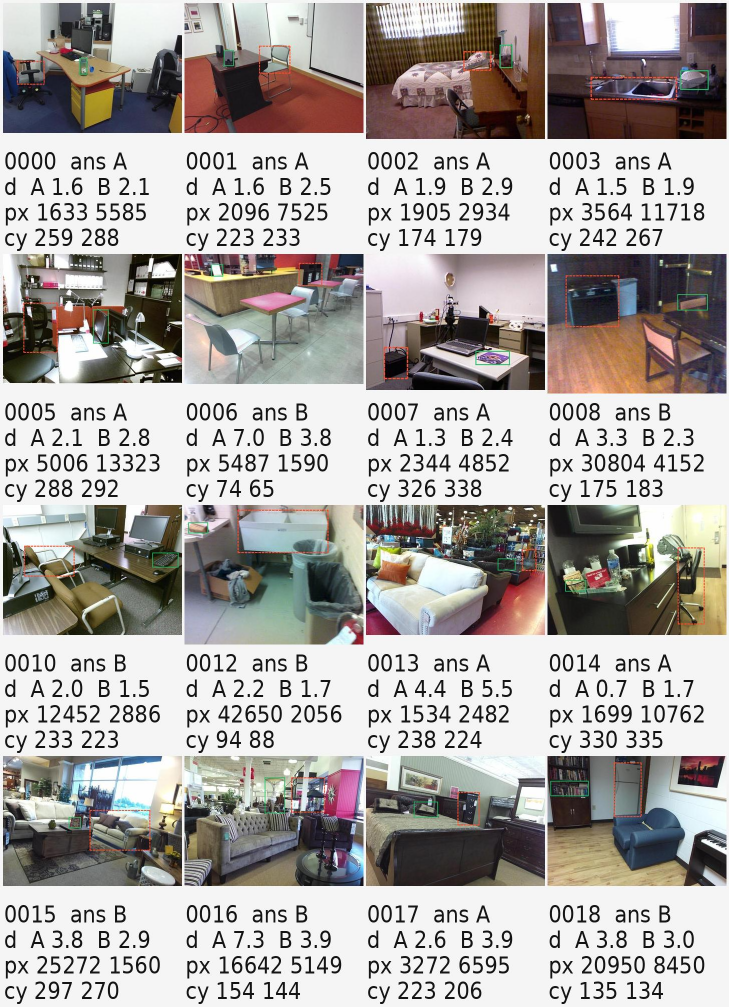}
  \caption{Size conflict}
\end{subfigure}\hfill
\begin{subfigure}[t]{0.48\linewidth}
  \centering
  \includegraphics[width=\linewidth,height=0.38\textheight,keepaspectratio]{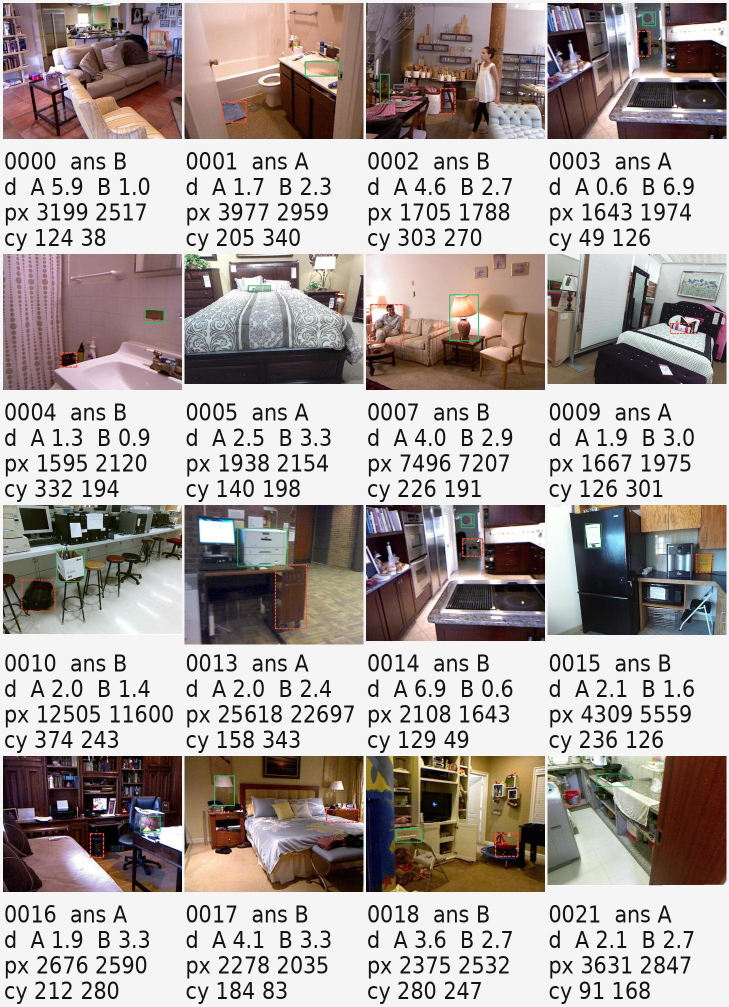}
  \caption{Vertical conflict}
\end{subfigure}
\caption{SUN RGB-D examples. Depth labels come from sensor depth inside
the object masks; cue conditions use measured mask area and centroid row.}
\label{fig:stimuli_real}
\end{figure}
\clearpage

%% file: appendix/a_protocols.tex
\section{Representation analysis and intervention}
\label{app:protocols}

\subsection{Probe fitting}
We fit logistic regression using mean binary cross-entropy plus
$10^{-3}\|w\|_2^2$, leaving the intercept unpenalized. This corresponds
to $\lambda=2\times10^{-3}$ under the convention in Eq.~\ref{eq:depth_probe}. Regularization is fixed across models
and domains rather than selected by search. Features are standardized
using training-fold statistics, with standard deviations floored at
$10^{-6}$. Optimization uses full-batch L-BFGS for at most 200 iterations
from zero initialization. Probe folds use an 80/20 random split;
the deployed direction uses seed 0.

An object token is selected when at least one quarter of its grid cell is
covered by the segmentation mask. If no cell passes this threshold, we
use the cell with the largest overlap. Object representations are
unweighted means of the selected tokens.

At each layer, the probe predicts depth order from $d=c_A-c_B$.
Standardization is fitted on the training fold only. The classifier uses
the standardized features and intercept in Eq.~\ref{eq:depth_probability};
the intervention uses the unit raw-space direction $v$ and the
offset-free sign of $v^\top d$. Classifier accuracy, sign accuracy, and
answer accuracy are therefore different measurements. Probes are fitted
on decorrelated calibration scenes. For real-image interventions, $v$ is
refitted on separate development images using Depth Anything~3
pseudo-depth~\citep{lin2025depth}. Evaluation uses sensor depth.

The category-only baseline predicts the majority depth ordering for
each unordered category pair observed in the probe's training fold.
Unseen category pairs receive a score of 0.5. On SUN RGB-D, this
baseline covers 22 of 209 size-conflict pairs and 8 of 144
vertical-conflict pairs. We therefore also evaluate the depth readout
on unseen category combinations. On the held-out probe fold,
these account for approximately 92\% of pairs, and Qwen3-VL-8B
achieves 84.4\% sign accuracy on them versus 84.5\% overall,
using the fixed intervention layer ($\ell=8$). Grey-image and mismatched-image
controls test image dependence. Fixed masks leave position information
in the grey-image control, and patch shuffling preserves local fog and
shading.
\begin{table}[htbp]
\centering
\small
\caption{Depth sign accuracy (\%) under image replacement.
The depth direction, question, and object-token positions are fixed.
Gemma's rendered controls use layer 20; the main intervention uses
layer 9.}
\label{tab:image-controls}
\begin{tabular}{llrrr}
\toprule
Model / domain & Conflict & Original & Grey & Mismatched \\
\midrule
Qwen3-8B, Render ($\ell=7$)
 & Size     & 99 & 52 & 49 \\
 & Vertical & 99 & 63 & 49 \\
Qwen3-8B, Real ($\ell=8$)
 & Size     & 84 & 51 & 58 \\
 & Vertical & 82.6 & 40.3 & 40 \\
Gemma-12B, Render ($\ell=20$)
 & Size     & 89 & 57 & 52 \\
 & Vertical & 99 & 32 & 52 \\
Gemma-12B, Real ($\ell=9$)
 & Size     & 83 & 57 & 55 \\
 & Vertical & 83 & 41 & 41 \\
\bottomrule
\end{tabular}
\end{table}

\subsection{Intervention and Setting Selection}

We apply object-level write-back once at the output of decoder
block $\ell-1$, where hidden-state layer 0 denotes the embeddings.
The edit occurs in the same forward pass used to compute the answer
logits.

Let $O_A$ and $O_B$ denote the queried objects' visual-token sets,
selected using the mask-overlap rule in Appendix~E.1. Before the edit,
we mean-pool their hidden states into $c_A$ and $c_B$ and compute
\[
z = v^\top(c_A-c_B),
\]
where $v$ is the unit-normalized depth direction in the original
hidden-state coordinates. It is obtained by dividing the fitted probe
weights by the feature standard deviations and normalizing; neither
the classifier's centering offset nor its intercept is used.

For each input, the update is
\[
\begin{aligned}
h_s' &= h_s + \alpha m\,\operatorname{sign}(z)
\bigl(\mathbb{1}[s\in O_A]-\mathbb{1}[s\in O_B]\bigr)v,\\
m &= \frac{1}{S}\sum_{u=1}^{S}\|h_u\|_2.
\end{aligned}
\]
Both $z$ and $m$ are computed from the pre-intervention states of
that input at layer $\ell$. The norm average includes all positions
in the left-padded sequence, so its value can depend on batch padding.
Only the queried object tokens are updated; shared tokens receive
zero net update. We use $\operatorname{sign}(0)=0$.

For disjoint object-token sets, the update changes the readout to
$z'=z+2\alpha m\,\operatorname{sign}(z)$, strengthening the depth
ordering already read from the model rather than supplying an
external ordering. Subsequent decoder blocks process the edited
states normally. The intervention uses no projection, gating,
or test-time refitting.

We select the layer $\ell$ and strength $\alpha$ on development data
by maximizing the smaller recovery of the two conflict gaps, subject
to at most a two-percentage-point loss on cue-consistent examples.
The direction and intervention settings are frozen before evaluation.
For Qwen3-VL-8B, the rendered experiments use $\ell=7$ and $\alpha=1.0$,
while SUN RGB-D uses $\ell=8$ and $\alpha=0.8$.

\begin{table}[htbp]
\centering
\small
\caption{Accuracy changes after write-back on rendered scenes
($n=500$ per split), reported as $\Delta$ [95\% CI]
in percentage points.}
\label{tab:render-ci}
\begin{tabular}{lcc}
\toprule
Model & Size conflict & Vertical conflict \\
\midrule
Qwen3-VL-4B
 & $+34.0\ [29.8,\,38.2]$ & $+40.6\ [36.2,\,45.0]$ \\
Qwen3-VL-8B
 & $+40.6\ [36.4,\,44.8]$ & $+28.6\ [24.8,\,32.6]$ \\
Qwen3-VL-32B
 & $+50.4\ [46.0,\,54.6]$ & $+39.4\ [35.2,\,43.6]$ \\
LLaVA-OV-1.5-8B
 & $+41.6\ [37.0,\,46.0]$ & $+39.6\ [35.4,\,44.0]$ \\
LLaVA-OV-2-8B
 & $+33.8\ [29.4,\,38.0]$ & $+28.6\ [24.4,\,32.8]$ \\
InternVL3-8B
 & $+29.0\ [25.2,\,33.0]$ & $+25.4\ [21.6,\,29.2]$ \\
InternVL3-38B
 & $+5.8\ [2.2,\,9.2]$ & $+8.0\ [5.0,\,11.0]$ \\
Gemma-4-12B
 & $+8.6\ [4.8,\,12.4]$ & $+3.6\ [0.2,\,7.0]$ \\
Gemma-4-31B
 & $+20.8\ [17.2,\,24.6]$ & $+10.2\ [7.2,\,13.2]$ \\
\bottomrule
\end{tabular}
\end{table}

\begin{table}[htbp]
\centering
\small
\caption{Accuracy changes after write-back on SUN RGB-D,
reported as $\Delta$ [95\% CI] in percentage points.}
\label{tab:real-ci}
\begin{tabular}{lcc}
\toprule
Model & Size conflict & Vertical conflict \\
\midrule
Qwen3-VL-4B
 & $+3.3\ [2.0,\,7.9]$ & $+6.2\ [2.0,\,12.6]$ \\
Qwen3-VL-8B
 & $+1.4\ [0.3,\,6.7]$ & $+5.6\ [2.3,\,12.2]$ \\
Qwen3-VL-32B
 & $+0.5\ [-0.3,\,6.0]$ & $+8.3\ [2.8,\,14.1]$ \\
LLaVA-OV-1.5-8B
 & $+12.0\ [5.9,\,18.1]$ & $+15.3\ [8.3,\,22.3]$ \\
LLaVA-OV-2-8B
 & $+6.2\ [1.4,\,11.1]$ & $+4.9\ [0.6,\,11.0]$ \\
InternVL3-8B
 & $+1.9\ [0.5,\,3.9]$ & $-1.4\ [-2.5,\,0.0]$ \\
InternVL3-38B
 & $0.0\ [0.0,\,0.0]$ & $0.0\ [-3.5,\,3.5]$ \\
Gemma-4-12B
 & $-7.7\ [-13.3,\,-2.4]$ & $-1.4\ [-7.6,\,5.0]$ \\
Gemma-4-31B
 & $-6.7\ [-12.6,\,-1.0]$ & $-5.6\ [-11.3,\,0.0]$ \\
\bottomrule
\end{tabular}
\end{table}

\subsection{Comparison with other methods}
Table~\ref{tab:method-comparison} compares current Qwen3-VL-8B results
on the same rendered splits. CoT changes the prompt; SpatioRoute changes
prompt routing; VCD and DoLa are
contrastive decoders~\citep{leng2024mitigating, chuang2024dola}; depth
input supplies a predicted depth map; and mean difference uses a
class-mean direction instead of the fitted probe direction. Each row uses
its registered evaluation setting. The intervention is much stronger on
both conflict splits and preserves the consistent split.

\begin{table}[htbp]
\centering
\small
\caption{Qwen3-VL-8B answer accuracy (\%) on rendered scenes.}
\label{tab:method-comparison}
\begin{tabular}{lccc}
\toprule
Method & Size conflict & Vertical conflict & Consistent \\
\midrule
Vanilla & \pid{1863} & \pid{1864} & \pid{1862} \\
CoT & \pid{2012} & \pid{2013} & \pid{2211} \\
SpatioRoute & \pid{2016} & \pid{2017} & \pid{2213} \\
VCD & \pid{2018} & \pid{2019} & \pid{2214} \\
DoLa & \pid{2020} & \pid{2021} & \pid{2215} \\
Depth input & \pid{2022} & \pid{2023} & \pid{2216} \\
Mean difference & \pid{2267} & \pid{2268} & \pid{2269} \\
\textbf{Intervention (ours)} & \textbf{\pid{2162}} & \textbf{\pid{2163}} & \textbf{\pid{2161}} \\
\bottomrule
\end{tabular}
\end{table}

\subsection{Controls and limitations}
\begin{figure}[htbp]
\centering
\includegraphics[width=0.92\linewidth]{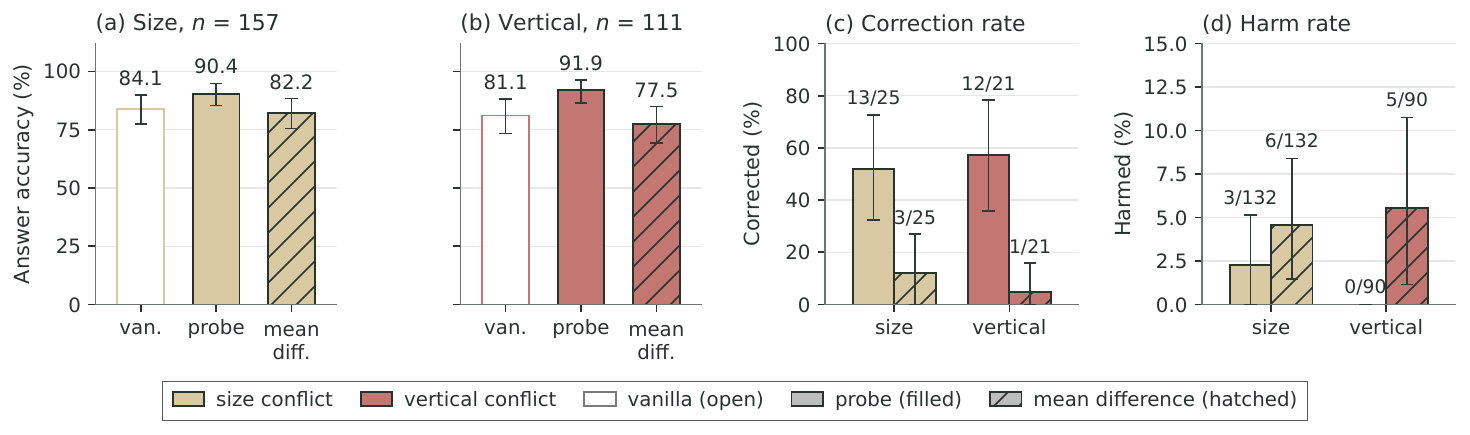}
\caption{Matched-readout comparison on real images. Size and vertical
subsets contain 157 and 111 pairs; labels give event counts.}
\label{fig:matched-readout}
\end{figure}

Reversing the direction reverses the behavioral effect. Random directions,
random signs, and writes to all image tokens or all tokens do not reproduce
the object-localized result. Figure~\ref{fig:matched-readout} compares the
probe direction with the mean-difference direction on SUN RGB-D pairs
where both read the correct depth order. With identical signs and
magnitudes, the probe direction corrects more initially wrong answers and
harms fewer initially correct answers. The 95\% intervals use an
image-cluster bootstrap.

The main-table effect varies by model and domain. It is weak for
InternVL3~\citep{zhu2025internvl3} on real images and harmful for
Gemma-4~\citep{team2026gemma}. The active-hook audit
for InternVL3-38B found nonzero readouts but no changed size-conflict
predictions; changes on the other real splits cancel in aggregate.
These are intervention failures, not evidence that the depth trace is
absent.